\documentclass{article}

\usepackage[preprint,nonatbib]{neurips_2021}

\usepackage[utf8]{inputenc} 
\usepackage[T1]{fontenc}    
\usepackage{hyperref}       
\usepackage{url}            
\usepackage{booktabs}       
\usepackage{amsfonts}       
\usepackage{nicefrac}       
\usepackage{microtype}      
\usepackage{xcolor}         

\usepackage{amsmath}
\usepackage{bbm}
\usepackage{comment}
\usepackage{graphicx}
\usepackage{aligned-overset}
\DeclareMathOperator*{\E}{\mathbb{E}}
\DeclareMathOperator*{\argmin}{argmin}

\DeclareMathOperator*{\B}{\mathbb{B}}
\DeclareMathOperator*{\scaleprod}{(\phi_{\omega}(s)-w_c)^{\top}(\phi_{\omega}(s)-w_c)}

\usepackage{algorithm}
\usepackage{algpseudocode}
\algnewcommand\algorithmicforeach{\textbf{for each}}
\algdef{S}[FOR]{ForEach}[1]{\algorithmicforeach\ #1\ \algorithmicdo}

\renewcommand{\eqref}[1]{eq.  \ref{#1}}
\newcommand{\secref}[1]{\S \ref{#1}}
\newcommand{\appref}[1]{Appendix \ref{#1}}
\newcommand{\metname}{DisTop}

\title{\metname: Discovering a Topological representation to learn diverse and rewarding skills}

\author{%
  Arthur Aubret, Laetitia Matignon, Salima Hassas  \\
    Univ Lyon, Université Lyon 1, CNRS, LIRIS F-69622\\
    Villeurbanne, France \\
    \texttt{\{firstname.lastname\}@univ-lyon1.fr} \\
}

\begin{document}

\maketitle

\begin{abstract}

The optimal way for a deep reinforcement learning (DRL) agent to explore is to learn a set of skills that achieves a uniform distribution of states. Following this, we introduce \metname, a new model that simultaneously learns diverse skills and focuses on improving rewarding skills. \metname~progressively builds a discrete topology of the environment using an unsupervised contrastive loss, a growing network and a goal-conditioned policy. Using this topology, a state-independent hierarchical policy can select where the agent has to keep discovering skills in the state space. In turn, the newly visited states allows an improved learnt representation and the learning loop continues. Our experiments emphasize that \metname~is agnostic to the ground state representation and that the agent can discover the topology of its environment whether the states are high-dimensional binary data, images, or proprioceptive inputs. We demonstrate that this paradigm is competitive on MuJoCo benchmarks with state-of-the-art algorithms on both single-task dense rewards and diverse skill discovery. By combining these two aspects, we show that \metname~achieves state-of-the-art performance in comparison with hierarchical reinforcement learning (HRL) when rewards are sparse. We believe \metname~opens new perspectives by showing that bottom-up skill discovery combined with representation learning can unlock the exploration challenge in DRL.



\end{abstract}

\section{Introduction}
In reinforcement learning (RL), an autonomous agent learns to solve a task by interacting with its environment \cite{sutton1998reinforcement} thus receiving a reward that has to be hand-engineered by an expert to efficiently guide the agent. However, when the reward is sparse or not available, the agent keeps having an erratic behavior. 

This issue partially motivated methods where an agent hierarchically  commits to a temporally extended behavior named skills \cite{dblpAubretMH20} or \textit{options} \cite{sutton1999between}, thus avoiding erratic behaviors and easing the exploration of the environment \cite{dblpMachadoBB17,aubret2019survey}. While it is possible to learn hierarchical skills with an extrinsic (\textit{i.e} task-specific) reward  \cite{bacon2017option,dblpRiemerLT18}, it does not fully address the exploration issue. In contrast, if an agent learns skills in a \textit{bottom-up} way with intrinsic rewards, it makes the skills task-independent \cite{aubret2019survey}: they are learnt without access to an extrinsic reward. It follows that, to explore without extrinsic rewards, the intrinsic objective of an agent may be to acquire a large and diverse repertoire of skills.  Such paradigm contrasts with the way prediction-based \cite{dblpBurdaEPSDE19} or count-based \cite{dblpBellemareSOSSM16} methods explore \cite{aubret2019survey} since their inability to return to the frontier of their knowledge can make their exploration collapse \cite{ecoffet2021first}. Previous approaches \cite{dblpNairPDBLL18,dblpPongDLNBL20,dblpEysenbachGIL19} manage to learn a large set of different skills, yet they are learnt during a developmental period \cite{metzen2013incremental}, which is just an unsupervised pre-training phase. 

These approaches based on a pre-training phase are incompatible with a \textit{continual learning} framework where the agent has to discover increasingly complex skills according to sequentially given tasks \cite{SilverYL13,parisi2019continual}. Ideally, a continually learning agent should learn new diverse skills by default, and focus on extrinsically rewarding skills when it gets extrinsic rewards \cite{pitis2020maximum,dblpAubretMH20}. It would make exploration very efficient since diverse skill learning can maximize the state entropy \cite{hazan2019provably,dblpPongDLNBL20}. To follow such principle, we assume that the agent should not focus on \textit{what} to learn, as commonly done in RL, but rather on \textbf{\textit{where} to learn in its environment.} Thus the agent should learn a representation that keeps the structure of the environment to be able to select where it is worth learning, instead of learning a representation that fits a wanted behavior which is most of the time unknown. 



In this paper, we introduce a new way to learn a goal space and select the goals, keeping the learning process end-to-end. We propose a new model that progressively \textbf{Dis}covers a \textbf{Top}ological representation (\textbf{\metname}) of the environment. \metname~bridges the gap between acquiring skills that reach a uniform distribution of embedded states, and  solving a task. It \textbf{makes diversity-based skill learning suitable for end-to-end exploration in a single-task setting irrespective of the ground state space}. \metname~simultaneously optimizes three components: 1- it learns a continuous representation of its states using a contrastive loss that brings together consecutive states; 2- this representation allows the building of a discrete topology of the environment with a new variation of a \textit{Growing When Required} network \cite{MarslandSN02}. Using the clusters of the topology, \metname~can sample from an almost-arbitrary distribution of visited embedded states, without approximating the high-dimensional ground state distribution; 3- it trains a goal-conditioned deep RL policy to reach embedded states. Upon these 3 components, a hierarchical state-independent Boltzmann policy selects the cluster of skills to improve according to an extrinsic reward and a diversity-based intrinsic reward.

We show, through \metname, that the paradigm of choosing \textit{where} to learn in a learnt topological representation is more generic than previous approaches. Our contribution is 4-folds: 1- we visualize the representation learnt by \metname~and exhibit that, unlike previous approaches \cite{dblpPongDLNBL20}, \metname~is agnostic to the shape of the ground state space and works with states being images, high-dimensional proprioceptive data or high-dimensional binary data; 2- we demonstrate that \metname~ clearly outperforms ELSIM, a recent method that shares similar properties \cite{dblpAubretMH20}; 3- we show that \metname~achieves similar performance with state-of-the-art (SOTA) algorithms on both single-task dense rewards settings \cite{haarnoja2018soft} and multi-skills learning benchmarks \cite{dblpPongDLNBL20}; 4- we show that it \textbf{improves exploration over state-of-the-art hierarchical methods} on hard hierarchical benchmarks.


\section{Background}\label{sec:background}\label{sec:skewfit}

\subsection{Goal-conditioned reinforcement learning for skill discovery}\label{sec:goalconditionedback}

In episodic RL, the problem is modelled with a Markov Decision Process (MDP). An agent can interact with an unknown environment in the following way: the agent gets a state $s_0 \in S$ that follows an initial state distribution $\rho_0(s)$. Its policy $\pi$ selects actions $a \in A$ to execute depending on its current state $s \in S$, $a_{execute}\sim \pi(a|s)$. Then, a new state is returned according to transition dynamics distribution $s'\sim p(s'|s,a)$. The agent repeats the procedure until it reaches a particular state or exceeds a fixed number of steps $T$. This loop is called an episode. An agent tries to adapt $\pi$ to maximize the expected cumulative discounted reward $\E_{\pi} \big[\sum_{t=0}^{T} \gamma^t R(s_{t-1},a_{t-1},s_t) \big]$ where $\gamma$ is the discount factor. The policy $\pi_{\theta}$ can be approximated with a neural network parameterized by $\theta$ \cite{haarnoja2018soft}.
In our case, the agent tries to learn \textit{skills}, defined as a policy that tries to target a state (which we name \textit{goal}). To this end, it autonomously computes an intrinsic reward according to the chosen skill \cite{aubret2019survey}. The intrinsic reward is approximated through parameters $\omega$. Following \textit{Universal Value Function Approximators} \cite{schaul2015universal}, the intrinsic reward  and the skill's policy become dependent on the chosen goal $g$: $r_t^g=R_{\omega}(s_{t-1},a_{t-1},s_t,g_t)$ and $\pi_{\theta}^g(s) = \pi_{\theta}(s,g)$. \textit{Hindsight Experience Replay} \cite{dblpAndrychowiczCRS17} is used to recompute the intrinsic reward to learn on arbitrary skills.


In particular, Skew-Fit \cite{dblpPongDLNBL20} strives to learn goal-conditioned policies that visit a maximal entropy of states. To generate goal-states with high entropy, they learn a generative model that incrementally increases the entropy of generated goals and makes visited states progressively tend towards a uniform distribution. Assuming a parameterized generative model $q_{\psi}(s)$ is available, an agent would like to learn $\psi$ to maximize the log-likelihood of the state distribution according to a uniform distribution: $\E_{s \sim U(S)} \log q_{\psi}(s)$. But sampling from the uniform distribution $U(S)$ is hard since the set of reachable states is unknown. Skew-Fit uses importance sampling and approximates the true ratio with a skewed distribution of the generative model $q_{\psi}(s)^{\alpha_{skew}}$ where $\alpha_{skew} < 0$. This way, it maximizes $\E_{s \sim Buffer} q_{\psi}(s)^{\alpha_{skew}} \log q_{\psi}(s)$ where $s$ are uniformly sampled from the buffer of the agent. $\alpha_{skew}$ defines the importance given to low-density states; if $\alpha_{skew}=0$, the distribution will stay similar, if $\alpha_{skew}=-1$, it strongly over-weights low-density states to approximate $U(S)$. In practice, they show that directly sampling states  $s\sim \frac{1}{Z}q_{\psi}(s)^{\alpha_{skew}}$, with $Z$ being the normalization constant reduces the variance.





\subsection{Contrastive learning and InfoNCE}\label{sec:infonce}

Contrastive learning aims at discovering a similarity metric between data points \cite{chopra2005learning}. First, positive pairs that represent similar data and fake negative pairs are built. Second they are given to an algorithm that computes a data representation able to discriminate the positive from the negative ones. InfoNCE \cite{dblpjournals/corr/abs-1807-03748} proposed to build positive pairs from states that belong to the same trajectory. The negative pairs are built with states randomly sampled. For a sampled trajectory, they build a set $S_N$ that concatenates $N-1$ negative states and a positive one $s_{t+p}$. Then, they maximize $\mathbb{L}_{InfoNCE}^N = -\E_{S_N} \big[ \log \frac{f_p(s_{t+p},v_t)}{\sum_{s_j \in S_N} f_p(s_j,v_t)} \big]$ where $v_t$ is a learnt aggregation of previous states, $p$ indicates the additional number of steps and $f_p$ is a positive similarity function. The numerator brings together states that are part of the same trajectory and the denominator pushes away negative pairs. InfoNCE maximizes a lower bound of the mutual information $I(c_t,s_{t+k})$ where $I(X,Y)= D_{KL}(p(x,y) || p(x)p(y))$ defines the quantity of information that a random variable contains on another. 



\section{Method}\label{sec:method}


We propose to redefine the interest of an agent so that its default behavior is to learn skills that achieve a uniform distribution of states while focusing its skill discovery in extrinsically rewarding areas when a task is provided. 
If the agent understands the true intrinsic environment structure and can execute the skills that navigate in the environment, a hierarchical policy would just have to select the state to target and to call for the corresponding skill. However, it is difficult to learn both skills and the structure. First the ground representation of states may give no clue of this structure and may be high-dimensional. Second, the agent has to autonomously discover new reachable states. Third, the agent must be able to easily sample goals where it wants to learn.

\begin{figure}[t]
    \centering
    \includegraphics[width=0.8\linewidth]{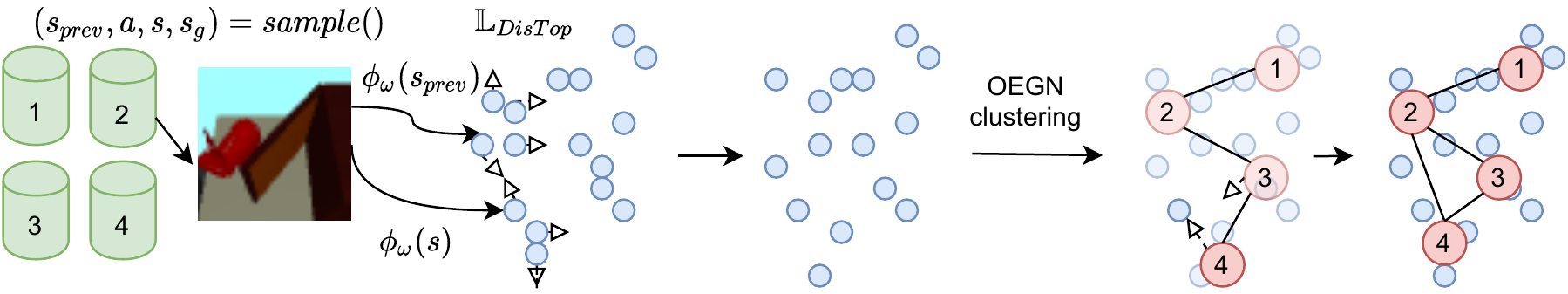}
    \caption{Illustration of a learning step of our growing network and contrastive loss (cf. text). Cylinders are buffers associated to each cluster, blue circles are states embedded with $\phi$ and pink circles represent clusters. The image is an example of state in the \textit{Visual Door}\cite{dblpNairPDBLL18} environment.}
    \label{fig:learningphase}
\end{figure}

 \metname~tackles theses challenges by executing skills that target low-density or previously rewarding visited states in a learnt topological representation. \figureautorefname~\ref{fig:learningphase} illustrates how \metname~learns a topological representation. It samples an interaction $(s_{prev},a,s,s_g)$ from different buffers (see below), where $s_g$ is the original goal-state, and embeds the associated observations with a neural network $\phi_{\omega}$. Then, $\phi_{\omega}$ is trained with the contrastive loss that guarantees the proximity between consecutive states (cf. \secref{sec:contrastive}). After that, our growing network dynamically clusters the embedded states in order to uniformly cover the embedded state distribution. \metname~approximates the probability density of visiting a state with the probability of visiting its cluster (cf. \secref{sec:topology}). Then, it assigns buffers to clusters, and can sample goal-states or states with almost-arbitrary density over the support of visited states, \textit{e.g} the uniform or reward-weighted ones (cf.  \secref{sec:sampling}). 
 
 At the beginning of each episode, a state-independent Boltzmann policy $\pi_{high}$ selects a cluster; then a state $s$ is selected that belongs to the buffer of the selected cluster and finally compute its representation $g=\phi_{\omega'}(s)$ (see \secref{sec:sampling}). A goal-conditioned policy $\pi_{\theta}^g$, trained to reach $g$, acts in the environment and discovers new reachable states close to its goal. The interactions made by the policy are stored in a buffer according to their initial objective and the embedding of the reached state.

\subsection{Learning a representation with a metric}\label{sec:contrastive}

\metname~strives to learn a state representation $\phi_{\omega}$ that reflects the topology of its environment. We propose to maximize the constrained mutual information between the consecutive states resulting from the interactions of an agent. To do this, \metname~takes advantage of the InfoNCE loss (cf. \secref{sec:infonce}). In contrast with previous approaches \cite{dblpjournals/corr/abs-1912-13414,dblpjournals/corr/abs-1807-03748,li2021learning}, we do not consider the whole sequence of interactions since this makes it more dependent on the policy. Typically, a constant and deterministic policy would lose the structure of the environment and could emerge once the agent has converged to an optimal deterministic policy. \metname~considers its \textit{local} variant and builds positive pairs with only two consecutive states. This way, it keeps distinct the states that cannot be consecutive and it more accurately matches the topology of the environment. We propose to select our similarity function as $f_{\omega}(s_t,s_{t+1})=e^{-k ||\phi_{\omega}(s_t)-\phi_{\omega}(s_{t+1})||_2}$ where $\phi_{\omega}$ is a neural network parameterized by $\omega$, $||_2$ is the L2 norm and $k$ is a temperature hyper-parameter \cite{chen2020simple}. If $\B$ is a batch of $N$ different consecutive pairs, the local InfoNCE objective, LInfoNCE, is described by \eqref{eq:boundedopt} (cf. \appref{app:deriveconstrastloss}).


\begin{align}
    \mathbb{L}_{LInfoNCE} &= \E_{(s_t,s_{t+1}) \in \B} \big[ \log 
    \frac{f_{\omega}(s_t,s_{t+1})
    }{\sum_{s \in \B}
    f_{\omega}(s_t,s)}
     \big]\nonumber \\
&\geq
    \E_{(s_t,s_{t+1})\in \B} \big[-k||\phi_{\omega}(s_t)-\phi_{\omega}(s_{t+1})||_2 
    - \log (
    1 + \sum_{s \in \B_{s \neq s_{t+1}}} f_{\omega}(s,s_{t+1}))
     \big] \label{eq:boundedopt}
\end{align}

We introduce the lower bound since it stabilizes the learning process. Intuitively, \eqref{eq:boundedopt} brings together states that are consecutive, and pushes away states that are separated by a large number of actions. This is illustrated in the second and third steps of \figureautorefname~\ref{fig:learningphase}. There are several drawbacks with this objective function: the representation may be strongly distorted, making a clustering algorithm inefficient, and may be too unstable to be used by a DRL algorithm. To tackle this, \eqref{eq:constrained} reformulates the objective as a constrained maximization. Firstly, \metname~guarantees that the distance between consecutive states is below a threshold $\delta$ (first constraint of \eqref{eq:constrained}). Secondly, it forces our representation to stay consistent over time (second constraint of \eqref{eq:constrained}). Weights $\omega'$ are a slow exponential moving average of $\omega$. 

\begin{equation}
\begin{aligned}
\max_{\omega} \quad \E_{(s_{t+1})\in \B}- \log (
    1 + \sum_{s \in \B} f_{\omega}(s,s_{t+1}) ) \quad
\textrm{s.t.} & \quad \E_{(s_t,s_{t+1})\in \B} ||\phi_{\omega}(s_t)-\phi_{\omega}(s_{t+1})||_2 \geq \delta \\
& \E_{s_{t+1}\in \B} ||\phi_{\omega}(s_{t+1})-\phi_{\omega'}(s_{t+1})||_2^2 = 0 \label{eq:constrained}
\end{aligned}
\end{equation}

Transforming the constraints into penalties, we get our final objective in \eqref{eq:objectiveloss}. $k_c$ is the temperature hyper-parameter that brings closer consecutive states, $\beta$ is the coefficient that slows down the speed of change of the representation.

\begin{align}
    \mathbb{L}_{\metname~}&=\E_{(s_t,s_{t+1})\in \B} \big[-k_c(relu(||\phi_{\omega}(s_t)-\phi_{\omega}(s_{t+1})||_2-\delta)) 
    - \log (
    1 + \sum_{s \in \B} f_{\omega}(s,s_{t+1}))
     \nonumber \\
     &+
     \beta ||\phi_{\omega}(s_{t+1})-\phi_{\omega'}(s_{t+1})||_2^2\big] \label{eq:objectiveloss}
\end{align}

By applying this objective function, \metname~learns a consistent, not distorted representation that keeps close consecutive states while avoiding collapse (see \figureautorefname~\ref{fig:gridworld} for an example). In fact, by increasing $k_c$ and/or $k$, one can select the level of distortion of the representation (cf. \appref{app:hyperparamstudy} for an analysis). One can notice that the function depends on the distribution of consecutive states in $\B$; we experimentally found that using tuples $(s_t,s_{t+1})$ from sufficiently stochastic policy is enough to keep the representation stable. We discuss the distribution of states that feed $\B$ in \secref{sec:learngoal}. In the following, we will study how \metname~takes advantage of this specific representation to sample diverse or rewarding state-goals.




\subsection{Mapping the topology to efficiently sample rare states}\label{sec:topology}

In this section, we define the skewed distribution of goal-states, so that \metname~increases the entropy of visited states by over-sampling low-density goal-states. We propose to skew the sampling of visited states by partitioning the embedded space using a growing network; then we just need to weight the probability of sampling the partitions. It will promote sampling low-density goal-states (cf. \secref{sec:sampling}) and will rule sampling for
learning the goal-conditioned and the representation (cf. \secref{sec:learngoal}). We first give the working principle before explaining how we define the skewed distribution and sample from it.

\paragraph{Discovering the topology.}

Since our representation strives to avoid distortion by keeping close consecutive states, \metname~can match the topology of the embedded environment by clustering the embedded states. \metname~uses an adaptation of a GWQ (cf. \secref{app:growings}) called Off-policy Embedded Growing Network (OEGN). OEGN dynamically creates, moves and deletes clusters so that clusters generates a network that uniformly cover the whole set of embedded states, independently of their density; this is illustrated in the two last steps of \figureautorefname~\ref{fig:learningphase}. Each node (or cluster) $c \in C$ has a reference vector $w_c$ which represents its position in the input space. The update operators make sure that all embedded states $s$ are within a ball centered on the center of its cluster, \textit{i.e} $min_c \scaleprod \leq \delta_{new}$ where $\delta_{new}$ is a threshold for creating clusters. $\delta_{new}$ is particularly important since it is responsible of the granularity of the clustering: if it is low, we will have a lot of small clusters, else we will obtain few large clusters that badly approximate the topology. The algorithm works as follows: assuming a new low-density embedded state is discovered, there are two possibilities: 1- the balls currently overlap: OEGN progressively moves the clusters closer to the new state and reduces overlapping; 2- the clusters almost do not overlap and OEGN creates a new cluster next to the embedded state. A learnt discrete topology can be visualized at the far right of \figureautorefname~\ref{fig:gridworld}. Details about OEGN can be found in \secref{app:growings}.



\paragraph{Sampling from a skewed distribution}


To sample more often low-density embedded states, we assume that the density of a visited state is reflected by the marginal probability of its cluster. So we approximate the density of a state with the density parameterized by $w$, reference vector of $s$: $q_w(s) \approx \frac{count(c_s)}{\sum_{c'\in C} count(c')} $ where $count(c_s)$ is the number of states that belong to the cluster that contains $s$. Using this approximation, we skew the distribution very efficiently by first sampling a cluster with the probability given by  $p_{\alpha_{skew}}(c) =\frac{count(c)^{1+\alpha_{skew}}}{\sum_{c'\in C} count(c')^{1+\alpha_{skew}}}  $ where $\alpha_{skew}$ is the skewed parameter (cf. \secref{sec:skewfit}). 
Then we randomly sample a state that belongs to the cluster. 




While our approximation $q_w(s)$ decreases the quality of our skewed objective, it makes our algorithm very efficient: we associate a buffer to each cluster and only weight the distribution of clusters; \metname~does not weight each visited state \cite{dblpPongDLNBL20}, but only a limited set of clusters. In practice we can trade-off the bias and the instability of OEGN by decreasing $\delta_{new}$: the smaller the clusters are, the smaller the bias is, but the more unstable is OEGN and the sampling distribution. So, in contrast with Skew-Fit, we do not need to compute the approximated density of each state, we just need to keep states in the buffer of their closest node. In practice, we associate to each node a second buffer that takes in interactions if the corresponding node is the closest to the interaction's goal. This results in two sets of buffers: $\B^G$ and $B^S$ to sample state-goals and states, respectively, with the skewed distribution.

In the next sections, we will detail how we use this skewed distribution over $\B^G$ and $B^S$ for sampling low-density goal-states (\secref{sec:sampling}) and sampling learning interactions (\secref{sec:learngoal}).

\subsection{Selecting novel or rewarding skills}\label{sec:sampling}

It is not easy to sample new reachable goals. For instance, using an embedding space $\mathbb{R}^{10}$, the topology of the environment may only exploit a small part of it, making most of the goals pointless. 
Similarly to previous works \cite{dblpWarde-FarleyWKI19}, \metname~generates goals by sampling previously visited states. To sample the states, \metname~first samples a cluster, and then samples a state that belongs to the cluster.

\paragraph{Sampling a cluster:} To increase the entropy of states, \metname~samples goal-states with the skewed distribution defined in \secref{sec:topology} that can be reformulated as:
\begin{align}
     p_{\alpha_{skew}}(c)
    &= \frac{e^{(1+\alpha_{skew}) \log  count(c)}}{\sum_{c'\in C}e^{{(1+\alpha_{skew})\log count(c')}}}
    \label{eq:noveltyboltzmann}
\end{align}

It is equivalent to sampling with a simple Boltzmann policy $\pi^{high}$, using a novelty bonus reward $\log count(c)$ and a fixed temperature parameter $1+\alpha_{skew}$. In practice, we can use a different $\alpha_{skew}'$ than in \secref{sec:topology} to trade-off the importance of the novelty reward in comparison with an extrinsic reward (see below) or decrease the speed at which we gather low-density states \cite{dblpPongDLNBL20}.



We can add a second reward to modify our skewed policy $\pi^{high}$ so as to take into consideration extrinsic rewards. We associate to a cluster $c \in C$ a value $r_c$ that represents the mean average extrinsic rewards received by the skills associated to its goals :
\begin{equation}
r_c = \E_{s \in c, g=\phi_{\omega}(s)} \frac{1}{T}\sum_{t=0}^T
 \E_{a_t\sim \pi_{\theta}^g(\cdot|s_t), s_{t+1} \sim p(\cdot|s_t,a_t)} R(s_t,a_t,s_{t+1}).
 \end{equation}
 The extrinsic value of a cluster $R_c^{ext}$ is updated with an exponential moving average $R_c^{ext} = (1-\alpha_c)*R_c^{ext} + \alpha_c*r_c$ where $\alpha_c$ is the learning rate. To favours exploration, we can also update the extrinsic value of the cluster's neighbors with a slower learning rate (cf. \appref{app:relabel}). Our sampling rule can then be :
 
  \begin{align}
    \pi^{high}(c) &= \text{softmax}_{C}(t^{ext} R_c^{ext} + (1+\alpha_{skew}') \log count(c)) \label{eq:boltzmannsamp}
\end{align}

\paragraph{Sampling a state:} Once the cluster is selected, we keep exploring independently of the presence of extrinsic rewards. To approximate a uniform distribution over visited states that belongs to the cluster, \metname~samples a vector in the ball that surrounds the center of the cluster. It rejects the vector and samples another one if it does not belong to the cluster. Finally, it selects the closest embedded state to the vector.

\subsection{Learning goal-conditioned policies}\label{sec:learngoal}

We now briefly introduce the few mechanisms used to efficiently learn the goal-conditioned policies. 
Once the goal-state $s_g$ is selected, the agent embeds it as $g=\phi_{\omega'}(s_g)$. It uses the target weights $\omega'$ to make our goal-conditioned policy independent from fluctuations due to the learning process. In consequence, we avoid using a hand-engineered environment-specific scheduling \cite{dblpPongDLNBL20}. Our implementation of the goal-conditioned policy is trained with Soft Actor-Critic (SAC)\cite{haarnoja2018soft} and the reward is the L2 distance between the state and the goal in the learnt representation. 
In practice, our goal-conditioned policy  $\pi_{\theta}$ needs a uniform distribution of goals to avoid forgetting previously learnt skills. Our representation function $\phi_{\omega}$ requires a uniform distribution over visited states to quickly learn the representation of novel areas. In consequence, \metname~samples a cluster $c \sim p_{\alpha_{skew}}$ and randomly takes half of the interactions from $\B^G$ and the rest from $\B^S$. We can also sample a ratio of clusters with $\pi^{high}$ if we do not care about forgetting skills (cf. \appref{app:hyperparamstudy}). $\theta$, $\omega$ and $w$ are learnt over the same mini-batch and we relabel the goals extracted from $\B^S$ (cf. \appref{app:relabel}).

\section{Experiments}\label{sec:expriments}

Our experiments aim at emphasizing the genericity of \metname. Firstly, unlike generative models in the ground state space, \metname~learns the true topology of the environment even though the ground state representation lacks structure. Secondly, when it does not access rewards, the discovered skills are as diverse as a SOTA algorithm (Skew-Fit \cite{dblpPongDLNBL20}) on robotic arm environments using images. Thirdly, even though acquiring diverse skills may be useless, \metname~is still able to perform close to a SOTA algorithm (SAC \cite{haarnoja2018soft}) in MuJoCo environments \cite{todorov2012mujoco}. Finally, we compare \metname~with a SOTA hierarchical algorithm (LESSON \cite{li2021learning}) in a sparse reward version of an agent that navigates in a maze \cite{nachum2019data}. All curves are a smooth averaged over 5 seeds, completed with a shaded area that represent the mean +/- the standard deviation over seeds. Used hyper-parameters are described in \appref{app:hyperparam}, environments and evaluation protocols details are given in \appref{app:environment}. Videos and images are available in supplementary materials.

\begin{figure}[t]
    \centering
    \includegraphics[width=0.9\linewidth]{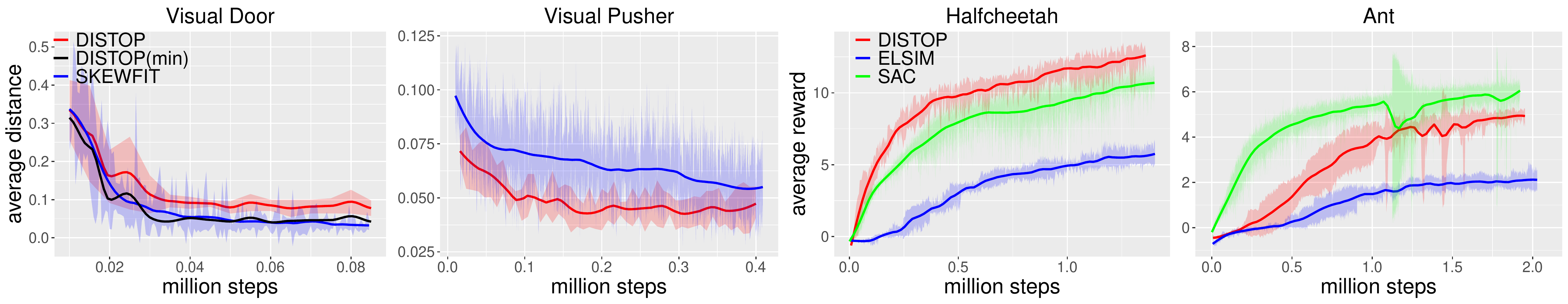}
    \caption{\textbf{Left:} Comparison of \metname~and Skew-Fit on their ability to reach diverse states. In the \textit{Visual Pusher} environment, we compare the final distance of the position puck with its desired position; in the door environment, we compare the angle of the door with the desired angle. DisTop(min) is the minimal distance reached through evaluation episode. At each evaluation iteration, the distances are averaged over fifty goals.
\textbf{Right:} Average rewards gathered throughout episodes of 300 steps while training on \textit{Halfcheetah-v2} and \textit{Ant-v2} environments. }
    \label{fig:mujoco}\label{fig:roboticv1}
\end{figure}

\paragraph{Is \metname~able to learn diverse skills without rewards ?} We assess the diversity of learnt skills on two robotic tasks \textit{Visual Pusher} (a robotic arm moves in 2D and eventually pushes a puck) and \textit{Visual Door} (a robotic arm moves in 3D and can open a door) \cite{dblpNairPDBLL18}. We compare to Skew-Fit \cite{dblpPongDLNBL20}. These environments are particularly challenging since the agent has to learn from 48x48 images using continuous actions. We use the same evaluation protocol than Skew-Fit: a set of images are sampled, given to the representation
function and the goal-conditioned policy executes the skill. 
In \figureautorefname~\ref{fig:roboticv1}, we observe that skills of \metname~are learnt quicker on \textit{Visual Pusher} but are slightly worst on \textit{Visual Door}. Since the reward function is similar, we hypothesize that this is due to the structure of the learnt goal space. In practice we observed that \metname~is more stochastic than Skew-Fit since the intrinsic reward function is required to be smooth over consecutive states.  Despite the stochasticity, it is able to reach the goal as evidenced by the minimal distance reached through the episode. Stochasticity does not bother evaluation on \textit{Visual Pusher} since the arm moves elsewhere after pushing the puck in the right position.

\paragraph{Does \metname~discover the environment topology even though the ground state space is unstructured ?}

\begin{figure}[t]
    \centering
    \includegraphics[width=0.9\linewidth]{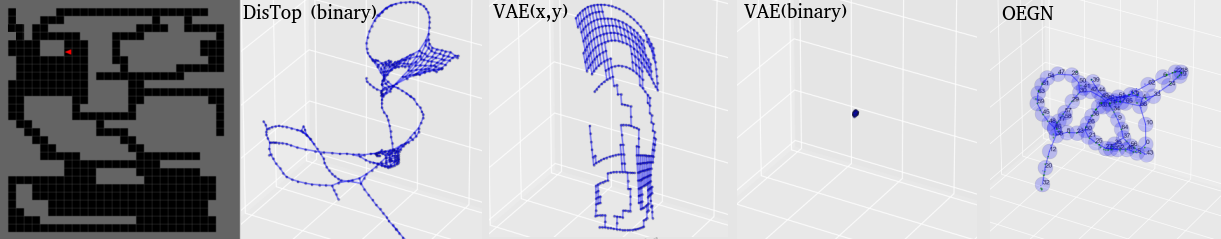}
    \caption{Visualization of the representations learnt by a VAE and \metname. From left to right, we respectively see a- the rendering of the maze; b- the representation learnt by \metname~with 900-dimensional binary inputs; c- a VAE representation with true (x,y) coordinates; d- a VAE representation with 900-dimensional binary inputs; e- OEGN network learnt from binary inputs.}
    \label{fig:gridworld}
\end{figure}

In this experiment, we analyze the representation learnt by a standard Variational Auto Encoder (VAE) and \metname. To make sure that the best representation is visualizable, we adapted the \textit{gym-minigrid} environment \cite{gym_minigrid} (cf. \figureautorefname~\ref{fig:gridworld}) where a randomly initialized agent moves for fifty timesteps in the four cardinal directions. Each observation is either a 900-dimensional one-hot vector with 1 at the position of the agent (\textit{binary}) or the (x,y) coordinates of the agent. Interactions are given to the representation learning algorithm. Except for OEGN, we display as a node the learnt representation of each possible state and connect states that are reachable in one step. In \figureautorefname~\ref{fig:gridworld}, we clearly see that \metname~is able to discover the topology of its environment since each connected points are distinct but close with each other. In contrast, the VAE representation collapses since it cannot take advantage of the proximity of states in the ground state space; it only learns a correct representation when it gets (x,y) positions, which are well-structured. This toy visualization highlights that, unlike VAE-based models, \metname~does not depend on well-structured ground state spaces to learn a suitable environment representation for learning skills.

\paragraph{Can \metname~solve non-hierarchical dense rewards tasks?} We test \metname~on MuJoCo environments 
\textit{Halfcheetah-v2} and \textit{Ant-v2} \cite{todorov2012mujoco}, the agent gets rewards to move forward as fast as possible. We fix the maximal number of timesteps to 300. We compare \metname~to our implementation of SAC\cite{haarnoja2018soft} and to ELSIM \cite{dblpAubretMH20}, a method that follows the same paradigm than \metname~(see \secref{sec:related}). 
We obtain better results than in the original ELSIM paper using similar hyper-parameters to \metname. In \figureautorefname~\ref{fig:mujoco} (Right), we observe that \metname~obtains high extrinsic rewards and clearly outperforms ELSIM. It also outperforms SAC on \textit{Halfcheetah-v2}, but SAC still learns faster on \textit{Ant-v2}. We highlight that SAC is specific to dense rewards settings and cannot learn in sparse settings (see below).



\begin{figure}[t]
    \centering
    \includegraphics[width=1\linewidth]{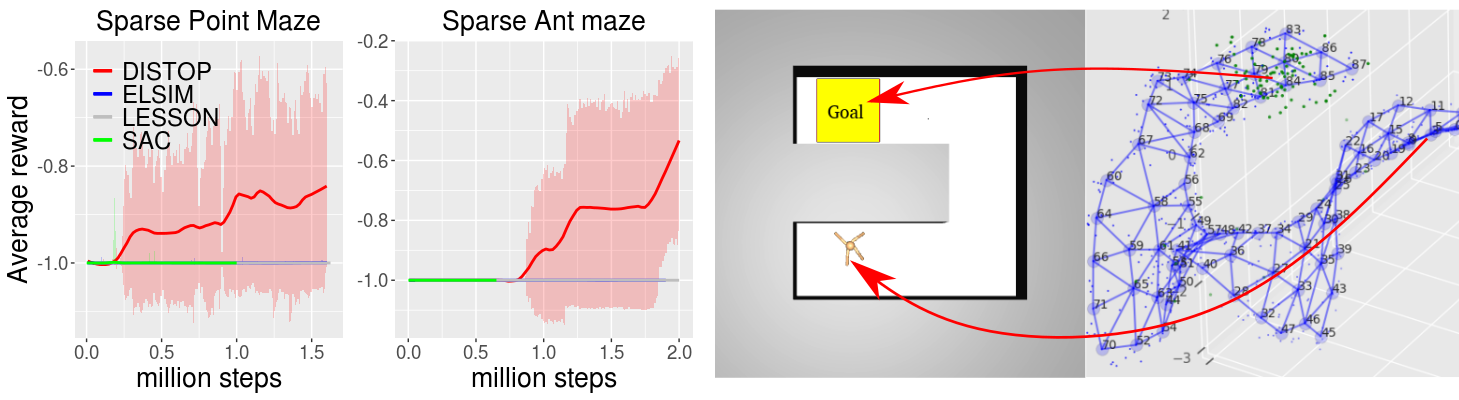}
    \caption{\textbf{Left:} Average rewards throughout training episodes. \textbf{Right:} Visualization of Ant Maze environment with its goal position and, to its right, an example of OEGN network learnt by \metname. Small green points represent selected goal-states.}
    \label{fig:mazev1}
\end{figure}

\paragraph{Does combining representation learning, entropy of states maximization and task-learning improve exploration on high-dimensional hierarchical tasks ?} We evaluate \metname~ability to explore and optimize rewards on image-based versions of \textit{Ant Maze} and \textit{Point Maze} environments \cite{nachum2019data}. The state is composed of a proprioceptive state of the agent and a top view of the maze ("Image"). In contrast with previous methods \cite{nachum2019data,li2021learning}, we remove the implicit \textit{curriculum} that changes the extrinsic goal across episodes; here we train only on the farthest goal and use a sparse reward function. Thus, the agent gets a non-negative reward only when it gets close to the goal. We compare our method with a SOTA HRL method, LESSON \cite{li2021learning} (see \secref{sec:related}), ELSIM and our implementation of SAC \cite{haarnoja2018soft}. For ELSIM, LESSON and \metname, we only pass the "image" to the representation learning part of the algorithms, assuming that an agent can separate its proprioceptive state from the sensorial state. In \figureautorefname~\ref{fig:mazev1}, we can see that \metname~is the only method that manages to return to the goal; in fact, looking at a learnt 3D OEGN network, we can see that it successfully represent the U-shape form of the maze and set goals close to the extrinsic goal. LESSON discovers the goal but does not learn to return to it\footnote{In \textit{Point Maze}, the best seed of LESSON returns to the goal after 5 millions timesteps}; we hypothesize that its high-level policy takes too long to gather enough interactions to learn on and that it still requires too many random high-level actions to find the goal. Neither SAC, nor ELSIM find the reward. We suppose that the undirected width expansion of the tree of ELSIM does not maximize the state-entropy, making it spend too much time in useless areas and thus inefficient for exploration. We admit that \metname~has high variance, but we argue that it is more a \textit{credit assignment} issue \cite{sutton1999between} due to the long-term reward than an exploration problem. 

\section{Related works}\label{sec:related}

        

\paragraph{Intrinsic skills in hierarchical RL}

Some works propose to learn hierarchical skills, but do not introduce a default behavior that maximizes the visited state entropy \cite{hazan2019provably}, limiting the ability of an agent to explore. For example, it is possible to learn skills that target a ground state or a change in the ground state space \cite{nachum2019data,levy2018hierarchical}. These approaches do not generalize well with high-dimensional states. To address this, one may want to generate rewards with a learnt representation of goals.
NOR \cite{nachum2018near} bounds the sub-optimality of such representation to solve a task and LESSON \cite{li2021learning} uses a slow dynamic heuristic to learn the representation. In fact, it uses an InfoNCE-like objective function; this is similar to \cite{dblpjournals/corr/abs-1912-13414} which learns the representation during pre-training with random walks. DCO \cite{jinnai2019exploration} generates \textit{options} by approximating the second eigenfunction of the combinatorial graph Laplacian of the MDP. It extends previous works \cite{dblpMachadoBB17,bar2020option} to continuous state spaces. Above-mentioned methods uses a hierarchical random walk to explore the environment, we have shown in \secref{sec:expriments} that \metname~explores quicker by maximizing the entropy of states in its topological representation.


\paragraph{Intrinsic motivation to learn diverse skills.} \metname~simultaneously learns skills, their goal representation, and which skill to train on. It contrasts with several methods that exclusively focus on selecting \textit{which skill to train on} assuming a good goal representation is available \cite{dblpFlorensaHGA18,dblpcurious,fournier2019clic,zhang2020automatic,dblpColasKLDMDO20}. They either select goals according to a curriculum defined with intermediate difficulty and the learning progress \cite{oudeyer2009intrinsic} or by imagining new language-based goals \cite{dblpColasKLDMDO20}.
In addition, \metname~strives to learn either skills that are diverse or extrinsically rewarding. It differs from a set of prior methods that learn only diverse skills during a pre-training phase, preventing exploration for end-to-end learning. Some of them maximize the mutual information (MI) between a set of states and skills. Typically, DIAYN \cite{dblpEysenbachGIL19}, VALOR \cite{dblpjournals/corr/abs-1807-10299} and SNN \cite{dblpFlorensaDA17} learn a discrete set of skills, but hardly generalize over skills. It has been further extended to continuous set of skills, using a generative model \cite{dblpSharmaGLKH20} or successor features \cite{dblpHansenDBWWM20,dblpBorsaBQMHMSS19}. In both case, directly maximizing this MI may incite the agent to focus only on simple skills \cite{campos2020explore}. DISCERN \cite{dblpWarde-FarleyWKI19} maximizes the MI between a skill and the last state of an episode using a contrastive loss. Unlike us, they use the true goal to generate positive pairs and use a L2 distance over pixels to define a strategy that improves the diversity of skills. In addition, unlike VAE-based models, our method better scales to any ground state space (see \secref{sec:expriments}). Typically, RIG \cite{dblpNairPDBLL18} uses a VAE \cite{dblpKingmaW13} to compute a goal representation before training the goal-conditioned policies. Using a VAE, it is possible to define a frontier with a reachability network, from which the agent should start stochastic exploration \cite{bharadhwaj2020leaf}; but the gradual extension of the frontier is not automatically discovered, unlike approaches that maximize the entropy of states (including \metname). Skew-Fit \cite{dblpPongDLNBL20} further extended RIG to improve the diversity of learnt skills by making the VAE over-weight low-density states. Unlike \metname, it is unclear how Skew-Fit could target another distribution over states than a uniform one. Approaches based on learning progress (LP) have already been built over VAEs \cite{kovavc2020grimgep,laversanne2018curiosity}; we believe that \metname~could make use of LP to avoid distractors or further improve skill selection.
\paragraph{Skill discovery for end-to-end exploration.} Like \metname, ELSIM \cite{dblpAubretMH20} can discover diverse and rewarding skills in an end-to-end way. It builds a tree of skills and selects the branch to improve with extrinsic rewards. 
\metname~outperforms ELSIM for both dense and sparse-rewards settings (cf. \secref{sec:expriments}). This end-to-end setting has also been experimented through multi-goal distribution matching \cite{pitis2020maximum,dblpleelisa} where the agent tries to reduce the difference between the density of visited states and a given distribution (with high-density in rewarding areas). Yet, either they approximate a distribution over the ground state space \cite{dblpleelisa} or assume a well-structured state representation \cite{pitis2020maximum}. Similar well-structured goal space is assumed when an agent maximizes the reward-weighted entropy of goals \cite{zhao2019maximum}. 
\paragraph{Dynamic-aware representations.} A set of RL methods try to learn a topological map without addressing the problem of discovering new and rewarding skills. Some methods \cite{dblpSavinovDK18,dblpSavinovRVMPLG19,DblpEysenbachSL19} consider a topological map over direct observations, but to give flat intrinsic rewards or make planning possible. We emphasize that SFA-GWR-HRL \cite{zhou2019vision} hierarchically takes advantage of a topological map built with two GWQ placed over two Slow Feature Analysis algorithms \cite{wiskott2002slow}; it is unclear whether it can be applied to other environments than their robotic setting. 
Functional dynamic-aware representations can be discovered by making the distance between two states match the expected difference of trajectories to go to the two states \cite{dblphoshGL19}; interestingly, they exhibit the interest of topological representations for HRL and propose to use a fix number of clusters to create goals. 
Previous work also showed that an active dynamic-aware search of independent factors can disentangle the controllable aspects of an environment \cite{dblpBengioTPPB17}. 


\section{Conclusion}\label{sec:conclusion}

We introduced a new model, \metname, that simultaneously learns a topology of its environment and the skills that navigate into it. In particular, it concentrates the skill discovery on parts of the topology with low-density embedded states until it discovers an extrinsic reward. In contrast with previous approaches, there is no pre-training \cite{pitis2020maximum,dblpPongDLNBL20}, particular scheduling \cite{dblpPongDLNBL20} or random walks \cite{dblpjournals/corr/abs-1912-13414,bharadhwaj2020leaf}. It avoids catastrophic forgetting, does not need a well-structured goal space \cite{pitis2020maximum} or dense rewards \cite{li2021learning}. 
Throughout all the experiments, we have shown that \metname~is generic : it keeps working and learn a dynamic-aware representation in dense rewards setting, pixel-based skill discovery environments and hard sparse rewards settings. Yet, there are limitations and exciting perspectives: HRL \cite{dblphoshGL19} and planning based approaches \cite{dblpNasirianyPLL19} could both take advantage of the topology and make easier states discovery; Frontier-based exploration \cite{bharadhwaj2020leaf} could also be explored to reduce skill stochasticity. Disentangling the topology \cite{bengio2013representation} and learning progress measures \cite{kovavc2020grimgep} could further improve the scalability of the approach. Finally, we argue that, by tackling the problem of unsupervised learning of both dynamic-aware representations and skills without adding constraints in comparison to standard RL, \metname~opens the path towards more human-like open-ended DRL agents.

\bibliographystyle{splncs}
\bibliography{references}

\begin{thebibliography}{10}
\providecommand{\url}[1]{\texttt{#1}}
\providecommand{\urlprefix}{URL }
\providecommand{\doi}[1]{https://doi.org/#1}

\bibitem{dblpjournals/corr/abs-1807-10299}
Achiam, J., Edwards, H., Amodei, D., Abbeel, P.: Variational option discovery
  algorithms. CoRR  \textbf{abs/1807.10299} (2018)

\bibitem{dblpAndrychowiczCRS17}
Andrychowicz, M., Crow, D., Ray, A., Schneider, J., Fong, R., Welinder, P.,
  McGrew, B., Tobin, J., Abbeel, P., Zaremba, W.: Hindsight experience replay.
  In: Annual Conference on Neural Information Processing Systems. pp.
  5048--5058 (2017)

\bibitem{aubret2019survey}
Aubret, A., Matignon, L., Hassas, S.: A survey on intrinsic motivation in
  reinforcement learning. arXiv preprint arXiv:1908.06976  (2019)

\bibitem{dblpAubretMH20}
Aubret, A., Matignon, L., Hassas, S.: {ELSIM:} end-to-end learning of reusable
  skills through intrinsic motivation. In: Hutter, F., Kersting, K., Lijffijt,
  J., Valera, I. (eds.) Machine Learning and Knowledge Discovery in Databases -
  European Conference, {ECML} {PKDD} 2020, Proceedings, Part {II}. Lecture
  Notes in Computer Science, vol. 12458, pp. 541--556. Springer (2020)

\bibitem{bacon2017option}
Bacon, P.L., Harb, J., Precup, D.: The option-critic architecture. In:
  Proceedings of the AAAI Conference on Artificial Intelligence. vol.~31 (2017)

\bibitem{bar2020option}
Bar, A., Talmon, R., Meir, R.: Option discovery in the absence of rewards with
  manifold analysis. In: International Conference on Machine Learning. pp.
  664--674. PMLR (2020)

\bibitem{dblpBellemareSOSSM16}
Bellemare, M.G., Srinivasan, S., Ostrovski, G., Schaul, T., Saxton, D., Munos,
  R.: Unifying count-based exploration and intrinsic motivation. In: Lee, D.D.,
  Sugiyama, M., von Luxburg, U., Guyon, I., Garnett, R. (eds.) Advances in
  Neural Information Processing Systems 29: Annual Conference on Neural
  Information Processing Systems 2016, December 5-10, 2016, Barcelona, Spain.
  pp. 1471--1479 (2016)

\bibitem{dblpBengioTPPB17}
Bengio, E., Thomas, V., Pineau, J., Precup, D., Bengio, Y.: Independently
  controllable features. CoRR  \textbf{abs/1703.07718} (2017)

\bibitem{bengio2013representation}
Bengio, Y., Courville, A., Vincent, P.: Representation learning: A review and
  new perspectives. IEEE transactions on pattern analysis and machine
  intelligence  \textbf{35}(8),  1798--1828 (2013)

\bibitem{bharadhwaj2020leaf}
Bharadhwaj, H., Garg, A., Shkurti, F.: Leaf: Latent exploration along the
  frontier. arXiv e-prints pp. arXiv--2005 (2020)

\bibitem{dblpBorsaBQMHMSS19}
Borsa, D., Barreto, A., Quan, J., Mankowitz, D.J., van Hasselt, H., Munos, R.,
  Silver, D., Schaul, T.: Universal successor features approximators. In: 7th
  International Conference on Learning Representations, {ICLR} 2019, New
  Orleans, LA, USA, May 6-9, 2019. OpenReview.net (2019)

\bibitem{dblpBurdaEPSDE19}
Burda, Y., Edwards, H., Pathak, D., Storkey, A.J., Darrell, T., Efros, A.A.:
  Large-scale study of curiosity-driven learning. In: 7th International
  Conference on Learning Representations, {ICLR} 2019, New Orleans, LA, USA,
  May 6-9, 2019. OpenReview.net (2019)

\bibitem{campos2020explore}
Campos, V., Trott, A., Xiong, C., Socher, R., Giro-i Nieto, X., Torres, J.:
  Explore, discover and learn: Unsupervised discovery of state-covering skills.
  In: International Conference on Machine Learning. pp. 1317--1327. PMLR (2020)

\bibitem{chen2020simple}
Chen, T., Kornblith, S., Norouzi, M., Hinton, G.: A simple framework for
  contrastive learning of visual representations. In: International conference
  on machine learning. pp. 1597--1607. PMLR (2020)

\bibitem{gym_minigrid}
Chevalier-Boisvert, M., Willems, L.: Minimalistic gridworld environment for
  openai gym. \url{https://github.com/maximecb/gym-minigrid} (2018)

\bibitem{chopra2005learning}
Chopra, S., Hadsell, R., LeCun, Y.: Learning a similarity metric
  discriminatively, with application to face verification. In: 2005 IEEE
  Computer Society Conference on Computer Vision and Pattern Recognition
  (CVPR'05). vol.~1, pp. 539--546. IEEE (2005)

\bibitem{dblpcurious}
Colas, C., Fournier, P., Sigaud, O., Oudeyer, P.: {CURIOUS:} intrinsically
  motivated multi-task, multi-goal reinforcement learning. CoRR
  \textbf{abs/1810.06284} (2018)

\bibitem{dblpColasKLDMDO20}
Colas, C., Karch, T., Lair, N., Dussoux, J., Moulin{-}Frier, C., Dominey, P.F.,
  Oudeyer, P.: Language as a cognitive tool to imagine goals in curiosity
  driven exploration. In: Larochelle, H., Ranzato, M., Hadsell, R., Balcan, M.,
  Lin, H. (eds.) Advances in Neural Information Processing Systems 33: Annual
  Conference on Neural Information Processing Systems 2020, NeurIPS 2020,
  December 6-12, 2020, virtual (2020)

\bibitem{ecoffet2021first}
Ecoffet, A., Huizinga, J., Lehman, J., Stanley, K.O., Clune, J.: First return,
  then explore. Nature  \textbf{590}(7847),  580--586 (2021)

\bibitem{DblpEysenbachSL19}
Eysenbach, B., Salakhutdinov, R., Levine, S.: Search on the replay buffer:
  Bridging planning and reinforcement learning. In: Wallach, H.M., Larochelle,
  H., Beygelzimer, A., d'Alch{\'{e}}{-}Buc, F., Fox, E.B., Garnett, R. (eds.)
  Advances in Neural Information Processing Systems 32: Annual Conference on
  Neural Information Processing Systems 2019, NeurIPS 2019, December 8-14,
  2019, Vancouver, BC, Canada. pp. 15220--15231 (2019)

\bibitem{dblpEysenbachGIL19}
Eysenbach, B., Gupta, A., Ibarz, J., Levine, S.: Diversity is all you need:
  Learning skills without a reward function. In: 7th International Conference
  on Learning Representations, {ICLR} 2019, New Orleans, LA, USA, May 6-9,
  2019. OpenReview.net (2019)

\bibitem{dblpFlorensaDA17}
Florensa, C., Duan, Y., Abbeel, P.: Stochastic neural networks for hierarchical
  reinforcement learning. In: 5th International Conference on Learning
  Representations, {ICLR} 2017, Toulon, France, April 24-26, 2017, Conference
  Track Proceedings. OpenReview.net (2017)

\bibitem{dblpFlorensaHGA18}
Florensa, C., Held, D., Geng, X., Abbeel, P.: Automatic goal generation for
  reinforcement learning agents. In: Dy, J.G., Krause, A. (eds.) Proceedings of
  the 35th International Conference on Machine Learning, {ICML} 2018,
  Stockholmsm{\"{a}}ssan, Stockholm, Sweden, July 10-15, 2018. Proceedings of
  Machine Learning Research, vol.~80, pp. 1514--1523. {PMLR} (2018)

\bibitem{fournier2019clic}
Fournier, P., Colas, C., Chetouani, M., Sigaud, O.: Clic: Curriculum learning
  and imitation for object control in non-rewarding environments. IEEE
  Transactions on Cognitive and Developmental Systems  (2019)

\bibitem{fritzke1995growing}
Fritzke, B., et~al.: A growing neural gas network learns topologies. Advances
  in neural information processing systems  \textbf{7},  625--632 (1995)

\bibitem{dblphoshGL19}
Ghosh, D., Gupta, A., Levine, S.: Learning actionable representations with goal
  conditioned policies. In: 7th International Conference on Learning
  Representations, {ICLR} 2019, New Orleans, LA, USA, May 6-9, 2019.
  OpenReview.net (2019), \url{https://openreview.net/forum?id=Hye9lnCct7}

\bibitem{haarnoja2018soft}
Haarnoja, T., Zhou, A., Abbeel, P., Levine, S.: Soft actor-critic: Off-policy
  maximum entropy deep reinforcement learning with a stochastic actor. In:
  International Conference on Machine Learning. pp. 1861--1870. PMLR (2018)

\bibitem{dblpHansenDBWWM20}
Hansen, S., Dabney, W., Barreto, A., Warde{-}Farley, D., de~Wiele, T.V., Mnih,
  V.: Fast task inference with variational intrinsic successor features. In:
  8th International Conference on Learning Representations, {ICLR} 2020, Addis
  Ababa, Ethiopia, April 26-30, 2020. OpenReview.net (2020)

\bibitem{hazan2019provably}
Hazan, E., Kakade, S., Singh, K., Van~Soest, A.: Provably efficient maximum
  entropy exploration. In: International Conference on Machine Learning. pp.
  2681--2691. PMLR (2019)

\bibitem{jinnai2019exploration}
Jinnai, Y., Park, J.W., Machado, M.C., Konidaris, G.: Exploration in
  reinforcement learning with deep covering options. In: International
  Conference on Learning Representations (2019)

\bibitem{dblpKingmaW13}
Kingma, D.P., Welling, M.: Auto-encoding variational bayes. In: Bengio, Y.,
  LeCun, Y. (eds.) 2nd International Conference on Learning Representations,
  {ICLR} 2014, Banff, AB, Canada, April 14-16, 2014, Conference Track
  Proceedings (2014)

\bibitem{kohonen1990self}
Kohonen, T.: The self-organizing map. Proceedings of the IEEE  \textbf{78}(9),
  1464--1480 (1990)

\bibitem{kovavc2020grimgep}
Kova{\v{c}}, G., Laversanne-Finot, A., Oudeyer, P.Y.: Grimgep: learning
  progress for robust goal sampling in visual deep reinforcement learning.
  arXiv preprint arXiv:2008.04388  (2020)

\bibitem{laversanne2018curiosity}
Laversanne-Finot, A., Pere, A., Oudeyer, P.Y.: Curiosity driven exploration of
  learned disentangled goal spaces. In: Conference on Robot Learning. pp.
  487--504. PMLR (2018)

\bibitem{dblpleelisa}
Lee, L., Eysenbach, B., Parisotto, E., Xing, E.P., Levine, S., Salakhutdinov,
  R.: Efficient exploration via state marginal matching. CoRR
  \textbf{abs/1906.05274} (2019)

\bibitem{levy2018hierarchical}
Levy, A., Platt, R., Saenko, K.: Hierarchical reinforcement learning with
  hindsight. In: International Conference on Learning Representations (2019)

\bibitem{li2021learning}
Li, S., Zheng, L., Wang, J., Zhang, C.: Learning subgoal representations with
  slow dynamics. In: International Conference on Learning Representations
  (2021), \url{https://openreview.net/forum?id=wxRwhSdORKG}

\bibitem{dblpjournals/corr/abs-1912-13414}
Lu, X., Tiomkin, S., Abbeel, P.: Predictive coding for boosting deep
  reinforcement learning with sparse rewards. CoRR  \textbf{abs/1912.13414}
  (2019)

\bibitem{dblpMachadoBB17}
Machado, M.C., Bellemare, M.G., Bowling, M.H.: A laplacian framework for option
  discovery in reinforcement learning. In: Precup, D., Teh, Y.W. (eds.)
  Proceedings of the 34th International Conference on Machine Learning, {ICML}
  2017, Sydney, NSW, Australia, 6-11 August 2017. Proceedings of Machine
  Learning Research, vol.~70, pp. 2295--2304. {PMLR} (2017)

\bibitem{MarslandSN02}
Marsland, S., Shapiro, J., Nehmzow, U.: A self-organising network that grows
  when required. Neural Networks  \textbf{15}(8-9),  1041--1058 (2002)

\bibitem{metzen2013incremental}
Metzen, J.H., Kirchner, F.: Incremental learning of skill collections based on
  intrinsic motivation. Frontiers in neurorobotics  \textbf{7}, ~11 (2013)

\bibitem{nachum2018near}
Nachum, O., Gu, S., Lee, H., Levine, S.: Near-optimal representation learning
  for hierarchical reinforcement learning. arXiv preprint arXiv:1810.01257
  (2018)

\bibitem{nachum2019data}
Nachum, O., Gu, S.S., Lee, H., Levine, S.: Data-efficient hierarchical
  reinforcement learning. In: Bengio, S., Wallach, H., Larochelle, H., Grauman,
  K., Cesa-Bianchi, N., Garnett, R. (eds.) Advances in Neural Information
  Processing Systems 31, pp. 3303--3313 (2018)

\bibitem{dblpNairPDBLL18}
Nair, A., Pong, V., Dalal, M., Bahl, S., Lin, S., Levine, S.: Visual
  reinforcement learning with imagined goals. In: Bengio, S., Wallach, H.M.,
  Larochelle, H., Grauman, K., Cesa{-}Bianchi, N., Garnett, R. (eds.) Advances
  in Neural Information Processing Systems 31: Annual Conference on Neural
  Information Processing Systems 2018, NeurIPS 2018, December 3-8, 2018,
  Montr{\'{e}}al, Canada. pp. 9209--9220 (2018)

\bibitem{dblpNasirianyPLL19}
Nasiriany, S., Pong, V., Lin, S., Levine, S.: Planning with goal-conditioned
  policies. In: Wallach, H.M., Larochelle, H., Beygelzimer, A.,
  d'Alch{\'{e}}{-}Buc, F., Fox, E.B., Garnett, R. (eds.) Advances in Neural
  Information Processing Systems 32: Annual Conference on Neural Information
  Processing Systems 2019, NeurIPS 2019, December 8-14, 2019, Vancouver, BC,
  Canada. pp. 14814--14825 (2019)

\bibitem{dblpjournals/corr/abs-1807-03748}
van~den Oord, A., Li, Y., Vinyals, O.: Representation learning with contrastive
  predictive coding. CoRR  \textbf{abs/1807.03748} (2018)

\bibitem{oudeyer2009intrinsic}
Oudeyer, P.Y., Kaplan, F.: What is intrinsic motivation? a typology of
  computational approaches. Frontiers in neurorobotics  \textbf{1}, ~6 (2009)

\bibitem{parisi2019continual}
Parisi, G.I., Kemker, R., Part, J.L., Kanan, C., Wermter, S.: Continual
  lifelong learning with neural networks: A review. Neural Networks
  \textbf{113},  54--71 (2019)

\bibitem{pitis2020maximum}
Pitis, S., Chan, H., Zhao, S., Stadie, B., Ba, J.: Maximum entropy gain
  exploration for long horizon multi-goal reinforcement learning. In:
  International Conference on Machine Learning. pp. 7750--7761. PMLR (2020)

\bibitem{dblpPongDLNBL20}
Pong, V., Dalal, M., Lin, S., Nair, A., Bahl, S., Levine, S.: Skew-fit:
  State-covering self-supervised reinforcement learning. In: Proceedings of the
  37th International Conference on Machine Learning, {ICML} 2020, 13-18 July
  2020, Virtual Event. Proceedings of Machine Learning Research, vol.~119, pp.
  7783--7792. {PMLR} (2020)

\bibitem{dblpRiemerLT18}
Riemer, M., Liu, M., Tesauro, G.: Learning abstract options. In: Bengio, S.,
  Wallach, H.M., Larochelle, H., Grauman, K., Cesa{-}Bianchi, N., Garnett, R.
  (eds.) Advances in Neural Information Processing Systems 31: Annual
  Conference on Neural Information Processing Systems 2018, NeurIPS 2018,
  December 3-8, 2018, Montr{\'{e}}al, Canada. pp. 10445--10455 (2018)

\bibitem{dblpSavinovDK18}
Savinov, N., Dosovitskiy, A., Koltun, V.: Semi-parametric topological memory
  for navigation. In: 6th International Conference on Learning Representations,
  {ICLR} 2018, Vancouver, BC, Canada, April 30 - May 3, 2018, Conference Track
  Proceedings. OpenReview.net (2018),
  \url{https://openreview.net/forum?id=SygwwGbRW}

\bibitem{dblpSavinovRVMPLG19}
Savinov, N., Raichuk, A., Vincent, D., Marinier, R., Pollefeys, M., Lillicrap,
  T.P., Gelly, S.: Episodic curiosity through reachability. In: 7th
  International Conference on Learning Representations, {ICLR} 2019, New
  Orleans, LA, USA, May 6-9, 2019. OpenReview.net (2019)

\bibitem{schaul2015universal}
Schaul, T., Horgan, D., Gregor, K., Silver, D.: Universal value function
  approximators. In: International conference on machine learning. pp.
  1312--1320. PMLR (2015)

\bibitem{dblpSharmaGLKH20}
Sharma, A., Gu, S., Levine, S., Kumar, V., Hausman, K.: Dynamics-aware
  unsupervised discovery of skills. In: 8th International Conference on
  Learning Representations, {ICLR} 2020, Addis Ababa, Ethiopia, April 26-30,
  2020. OpenReview.net (2020)

\bibitem{SilverYL13}
Silver, D.L., Yang, Q., Li, L.: Lifelong machine learning systems: Beyond
  learning algorithms. In: Lifelong Machine Learning, Papers from the 2013
  {AAAI} Spring Symposium, Palo Alto, California, USA, March 25-27, 2013.
  {AAAI} Technical Report, vol. {SS-13-05}. {AAAI} (2013)

\bibitem{sutton1998reinforcement}
Sutton, R.S., Barto, A.G.: Reinforcement learning: An introduction, vol.~1. MIT
  press Cambridge (1998)

\bibitem{sutton1999between}
Sutton, R.S., Precup, D., Singh, S.: Between mdps and semi-mdps: A framework
  for temporal abstraction in reinforcement learning. Artificial intelligence
  \textbf{112}(1-2),  181--211 (1999)

\bibitem{todorov2012mujoco}
Todorov, E., Erez, T., Tassa, Y.: Mujoco: A physics engine for model-based
  control. In: IEEE/RSJ {IROS}. pp. 5026--5033. IEEE (2012)

\bibitem{dblpWarde-FarleyWKI19}
Warde{-}Farley, D., de~Wiele, T.V., Kulkarni, T.D., Ionescu, C., Hansen, S.,
  Mnih, V.: Unsupervised control through non-parametric discriminative rewards.
  In: 7th International Conference on Learning Representations, {ICLR} 2019,
  New Orleans, LA, USA, May 6-9, 2019. OpenReview.net (2019)

\bibitem{wiskott2002slow}
Wiskott, L., Sejnowski, T.J.: Slow feature analysis: Unsupervised learning of
  invariances. Neural computation  \textbf{14}(4),  715--770 (2002)

\bibitem{zhang2020automatic}
Zhang, Y., Abbeel, P., Pinto, L.: Automatic curriculum learning through value
  disagreement. Advances in Neural Information Processing Systems  \textbf{33}
  (2020)

\bibitem{zhao2019maximum}
Zhao, R., Sun, X., Tresp, V.: Maximum entropy-regularized multi-goal
  reinforcement learning. In: International Conference on Machine Learning. pp.
  7553--7562. PMLR (2019)

\bibitem{zhou2019vision}
Zhou, X., Bai, T., Gao, Y., Han, Y.: Vision-based robot navigation through
  combining unsupervised learning and hierarchical reinforcement learning.
  Sensors  \textbf{19}(7), ~1576 (2019)

\end{thebibliography}

\newpage

\appendix

\section{Ablation study}\label{app:hyperparamstudy}

In this section, we study the impact of the different key hyper-parameters of \metname. Except for paper results, we average the results over 3 random seeds. For visualizations based on the topology, the protocol is the same as the one described \secref{sec:expriments}. In addition, we select the most viewable one among 3 learnt topologies; while we did not observe variance in our analysis through the seeds, the 3D angle of rotation can bother our perception of the topology.

\paragraph{Details about maze experiments.} 

\begin{figure}[t]
    \centering
    \includegraphics[width=0.8\linewidth]{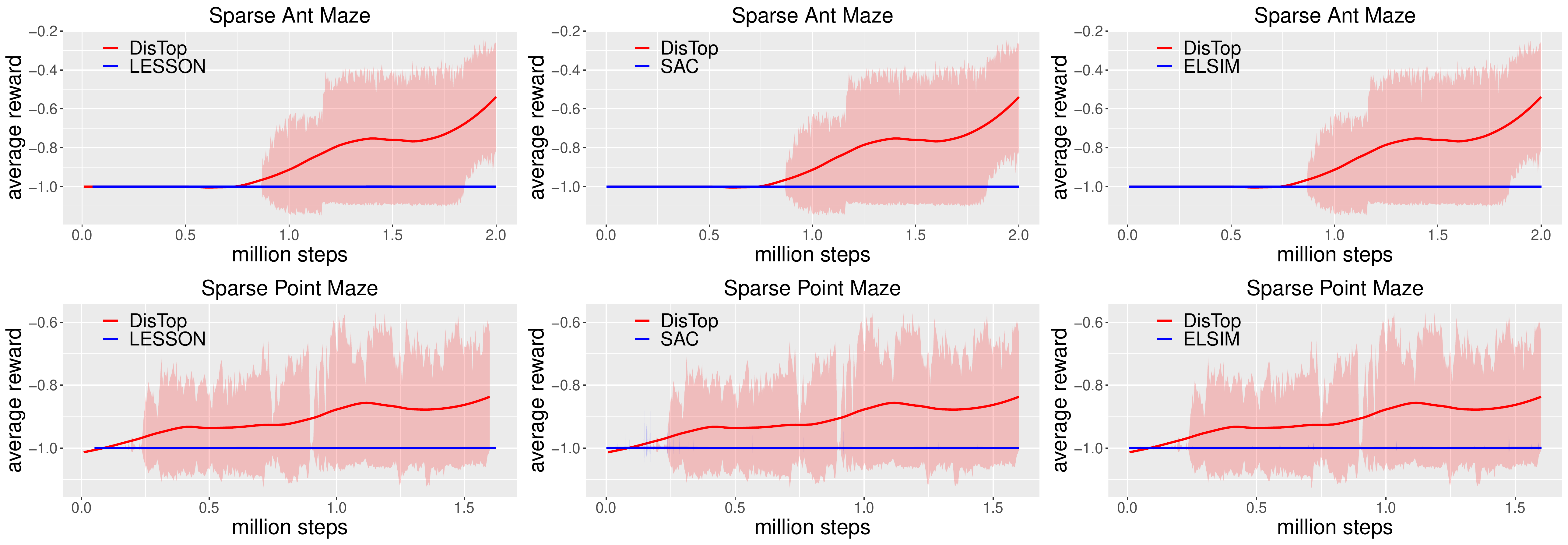}
    \caption{Same experiments than in \secref{sec:expriments}.} 
    \label{fig:mazeadd}
\end{figure}

\figureautorefname~\ref{fig:mazeadd} shows the same experiments as in \secref{sec:expriments}, but we duplicate the graph to clearly see the curves of each method. For SAC, we use different experiments with $\gamma=0.996$ rather than $\gamma=0.99$.

\paragraph{Controlling the distortion of the representation.} The analysis of our objective function $\mathbb{L}_{\metname~}$ shares some similarities with standard works on contrastive learning \cite{chopra2005learning}. However, we review it to clarify the role of the representation with respect to the interactions, the reward function and OEGN.

\begin{figure}[t]
    \centering
    \includegraphics[width=0.6\linewidth]{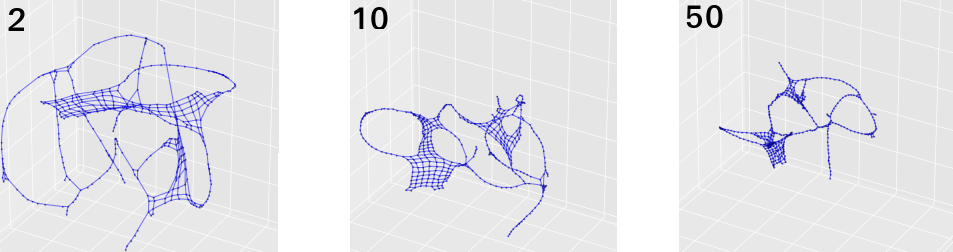}
    \caption{Different Topologies learnt on the gridworld displayed in \secref{sec:expriments}. From left to right, we show the learnt topology with $k_c=2$, $k_c=10$, $k_c=50$.}
    \label{fig:distort}
\end{figure}

\begin{figure}[t]
    \centering
    \includegraphics[width=1\linewidth]{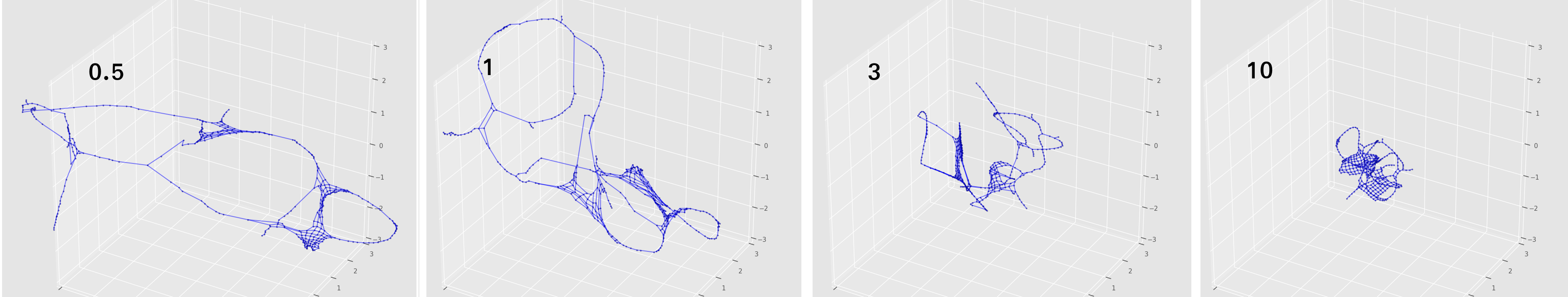}
    \caption{Different Topologies learnt on the gridworld displayed in \secref{sec:expriments}. From left to right, we show the learnt topology with $k=0.5$, $k=1$, $k=3$, $k=10$.}
    \label{fig:distorttau}
\end{figure}

In \figureautorefname~\ref{fig:distort}, we can see that the distortion parameter $k_c$ rules the global dilatation of the learnt representation. A low $k_c$ also increases the distortion of the representation, which may hurt the quality of the clustering algorithm. $k_c$ is competing with $k$, the temperature hyper-parameter. As we can see in \figureautorefname~\ref{fig:distorttau}, $k$ rules the minimal allowed distance between very different states. So that there is a trade-off between a low $k$ that distorts and makes the representation unstable, and a high $k$ that allows different states to be close with each other. With a high $k$, the L2 rewards may admit several local optimas.

\begin{figure}[t]
    \centering
    \includegraphics[width=1\linewidth]{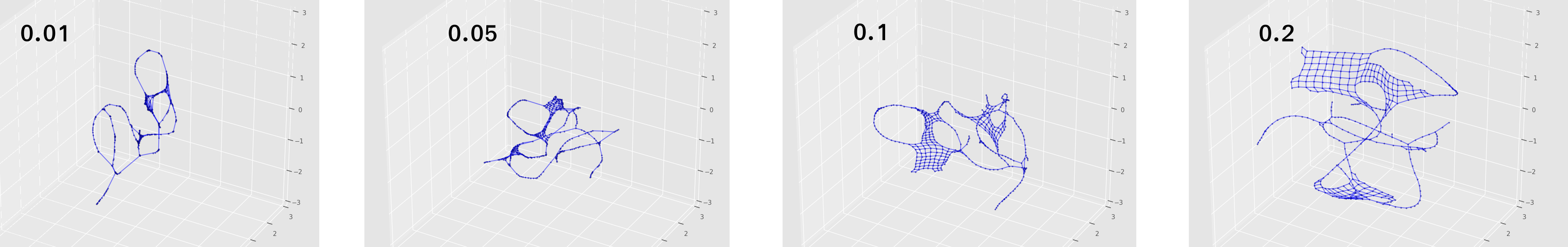}
    \caption{Different Topologies learnt on the gridworld displayed in \secref{sec:expriments}. From left to right, we show the learnt topology with $\delta=0.01$, $\delta=0.05$, $\delta=0.1$, $\delta=0.2$.}
    \label{fig:distortrp}
\end{figure}

In \figureautorefname~\ref{fig:distortrp}, we see that $\delta$ also impacts the distortion of the representation, however it mostly limits the compression of the representation in the areas the agent often interacts in. In the borders, the agent often hurts the wall and stays in the same position. So in comparison with large rooms, the \textit{bring together} is less important than the \textit{move away} part. $\delta$ limits such asymmetries and keeps large rooms dilated. Overall, this asymmetry also occurs when the number of states increases due to the exploration of the agent: the agent progressively compresses its representation since interesting negative samples are less frequent.

\begin{figure}[t]
    \centering
    \includegraphics[width=1\linewidth]{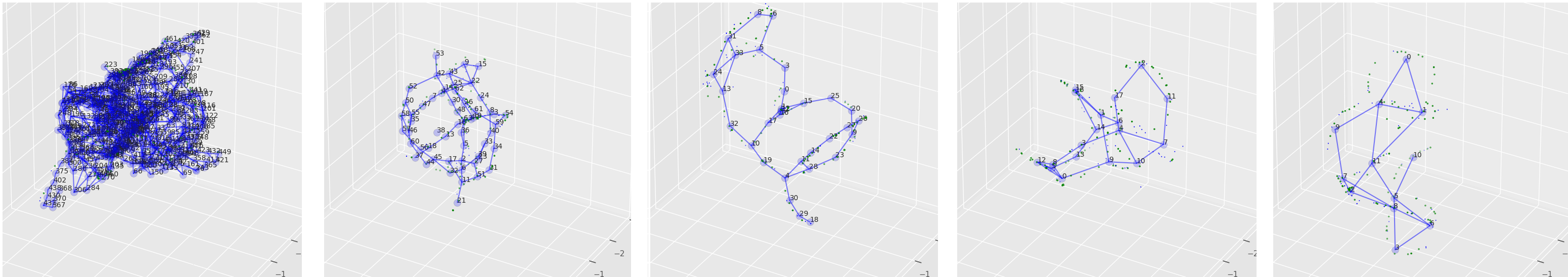}
    \caption{Different OEGN networks learnt according to $\delta_{new}$. From left to right, we show the OEGN network with $\delta_{new}=0.2$,$\delta_{new}=0.4$,$\delta_{new}=0.6$,$\delta_{new}=0.8$,$\delta_{new}=1$}
    \label{fig:oegnthres}
\end{figure}

The size of the clusters of OEGN has to match the distortion of the representation. \figureautorefname~\ref{fig:oegnthres} emphasizes the importance of the creation threshold parameter $\delta_{new}$. If this is too low ($0.2$), clusters do not move and a lot of clusters are created. OEGN waits a long time to delete them. If it is too high, the approximation of the topology becomes very rough, and states that belong to very different parts of the topology are classified in the same cluster; this may hurt our density approximation.

\paragraph{Selection of interactions:}

\begin{figure}[t]
    \centering
    \includegraphics[width=1\linewidth]{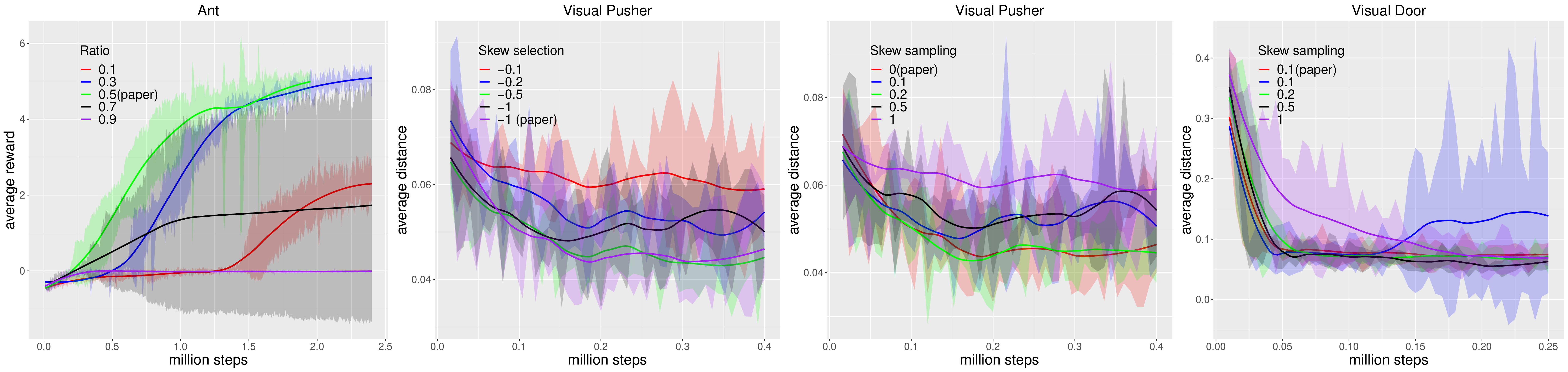}
    \caption{Different learning curves showing the impact of the choice of interactions. 1- We study the impact choosing learning interactions with $\pi^{high}$ rather than $p_{\alpha_{skew}}$. 2- we study the importance of $1+\alpha_{skew}'$ in \textit{Visual Pusher}. 3 and 4- we assess the importance of $1+\alpha_{skew}$ in \textit{Visual Pusher} and \textit{Visual Door}.
    }
    \label{fig:selection}
\end{figure}

\figureautorefname~\ref{fig:selection} shows the importance of the different hyper-parameters that rule the sampling of goals and states. In Ant environment, we see that the agent has to sample a small ratio of learning interactions from  $\pi^{high}$ rather than $p_{\alpha_{skew}}$; it speeds up its learning process by making it focus on important interactions. Otherwise, it is hard to learn all skills at the same time. However, the learning process becomes unstable if it deterministically samples from a very small set of clusters (ratios $0.9$ and $0.7$). 

Then we evaluate the importance of $1+\alpha_{skew}'$ on \textit{Visual Pusher}. We see that the agent learns quicker with a low $1+\alpha_{skew}'$. It hardly learns when the high-level policy almost does not over-sample low-density clusters ($-0.1$). This makes our results consistent with the analysis provided in the paper of Skew-Fit \cite{dblpPongDLNBL20}.

In the last two graphics of \figureautorefname~\ref{fig:selection}, we show the impact of $1+\alpha_{skew}$; we observe that the agent learns quicker when $1+\alpha_{skew}$ is close to $0$. It highlights that the agent quicker learns both a good representation and novel skills by sampling uniformly over the clusters. 

Overall, we also observe that some seeds become unstable. Since they coincide with large deletions of clusters. We expect that an hyper-parameter search on the delete operators of OEGN may solve this issue in these specific cases.

\section{OEGN and GWQ}\label{app:growings}

Several methods introduced unsupervised neural networks to learn a discrete representation of a fixed input distribution \cite{kohonen1990self,fritzke1995growing,MarslandSN02}. The objective is to have clusters (or nodes) that cover the space defined by the input points. Each node $c \in C$ has a reference vector $w_c$ which represents its position in the input space. We are interested in the \textit{Growing when required} algorithm (GWQ) \cite{MarslandSN02} since it updates the structure of the network (creation and deletion of edges and nodes) and does not impose a frequency of node addition. OEGN is an adaptation of GWQ; algorithm \ref{algo:oegn} describes the core of these algorithms and their specific operators are described below. Specific operations of OEGN are colored in red, the ones of GWQ in blue and the common parts are black. Our modifications are made to  1- make the network aware of the dynamics; 2- adapt the network to the dynamically changing true distribution.

\paragraph{Delete operator:} \textcolor{red}{Delete $node_i$ if $node_i.error$ is above a threshold $\delta_{error}$ to verify that the node is still active; delete the less filled neighbors of two neighbors if their distance is below $\delta_{prox}$ to avoid too much overlapping; check it has been selected $n_{del}$ times before deleting it.} Delete if a node does not have edges anymore.

\paragraph{Both creation and moving operators:} \textcolor{red}{Check that the distance between the original goal and the resulting embedding $||goal_i-e_i||_2$ is below a threshold $\delta_{success}$.}

\paragraph{Creation operator:} Check if $||closest-e_i||_2 > \delta_{new}$. \textcolor{red}{ Verify $closest$ has already been selected $\delta_{count}$ times by $\pi^{high}$. If the conditions are filled, a new node is created at the position of the incoming input $e_i$ and it is connected to $closest$.}
\textcolor{blue}{Update a firing count of the winning node and its neighbors and check it is below a threshold. A node is created halfway between the winning node and the second closest node. The two winning nodes are connected to the new node.}

\paragraph{Moving operator:} If no node is created or deleted, which happens most of the time, we apply the moving operator described by eq. \ref{eq:finalupdatein} assuming $closest=\argmin_c (\phi_{\omega'}(s)-w_c)^{\top}(\phi_{\omega'}(s)-w_c)$. In our case, we use a very low $\alpha_{neighbors}$ to avoid the nodes to concentrate close to high-density areas.

\begin{align}
    w_j = \begin{cases}
    w_j + \alpha
    (\phi_{\omega'}(s)-w_j), 
    & \text{if $j=closest$}\\
    w_j + \alpha_{neighbors}
    (\phi_{\omega'}(s)-w_j), 
    & \text{if $j \in neighbors(closest)$}\\
    w_j, & \text{otherwise.}
    \end{cases} \label{eq:finalupdatein}
\end{align}

\paragraph{Edge operator:} \textcolor{red}{edges are added or updated with attribute $age=0$ between the winning node of $e_i$ and the one of $e_i^{prev}$.} \textcolor{blue}{Edges with $age=0$ are added or updated between the two closest nodes of $e_i$.} When an edge is added or updated, increment the age of the other neighbors of $closest$ and delete edges that are above a threshold $\delta_{age}$.


\begin{algorithm}
\begin{algorithmic}
\State Initialize network with two random nodes, set theirs attributes to 0 and connect them.
\ForEach{learning iteration}

\State \textcolor{red}{Sample a tuple $s_i, node_i, s_i^{prev} \leftarrow sample(\B)$.}
\State \textcolor{red}{Embed states: $e_i \leftarrow \phi_{\omega'}(s_i)$; $e_i^{prev} \leftarrow \phi_{\omega'}(s_i^{prev})$}
\State \textcolor{blue}{Sample an input $e_i \leftarrow sample(\B)$.}
\State $closest \leftarrow min_{nodes}(||nodes-e_i||_2)$.
\State Increase error count of $node_i$ by 1.
\State Reset error count of $closest$ to 0
\State Apply $DeleteOperator()$.
\State If a node is deleted, stop the learning iteration
\State Apply $CreationOperator()$.
\State Apply $MovingOperator()$.
\State Apply $EdgeOperator()$.


\EndFor
\end{algorithmic}
\caption{Algorithm of OEGN (red) and GWQ (blue)}
\label{algo:oegn}
\end{algorithm}

Except for $\delta_{new}$ and $\delta_{success}$, we emphasize that thresholds parameters have not been fine-tuned and are common to all experiments; we give them in Table \ref{tab:oegnhype}.

\begin{table}[]
    \centering
    \begin{tabular}{|c|c|}
        \hline
         Parameters & Value \\\hline 
         Age deletion threshold $\delta_{age}$ & $600$ \\ 
        Error deletion count $\delta_{error}$ & $600$ \\ 
        Proximity deletion threshold $\delta_{prox}$ & $0.4 \times \delta_{new}$ \\ 
         Count creation threshold $\delta_{count}$ & $5$ \\ 
         Minimal number of selection $n_{del}$ & $10$ \\ 
         Learning rate $\alpha$ & $0.001$ \\ \hline
         Neighbors learning rate $\alpha_{neighbors}$ & $0.000001$ \\
         Number of updates per batch & $\sim 32$ \\\hline

    \end{tabular}
    \caption{Fixed hyper-parameters used in OEGN. They have not been tuned.}
    \label{tab:oegnhype}
\end{table}

\section{Derivation of the contrastive loss}\label{app:deriveconstrastloss}

\begin{align}
    \mathbb{L}_{InfoNCE} &= \E_{(s_t,s_{t+1}) \in \B} \big[ \log 
    \frac{f_{\omega}(s_t,s_{t+1})
    }{\sum_{s \in \B}
    f_{\omega}(s,s_{t+1})}
     \big]\nonumber \\
&=
    \E_{(s_t,s_{t+1})\in \B} \big[-k||\phi_{\omega}(s_t)-\phi_{\omega}(s_{t+1})||_2 
    - \log ( \sum_{s \in \B} f_{\omega}(s,s_{t+1}))
     \big] \\
&=
    \E_{(s_t,s_{t+1})\in \B} \big[-k||\phi_{\omega}(s_t)-\phi_{\omega}(s_{t+1})||_2 
    - \log (
    f_{\omega}(s_t,s_{t+1}) + 
    \sum_{s \in \B_{s \neq s_{t}}} f(s,s_{t+1}) )
     \big] \\
&\geq
    \E_{(s_t,s_{t+1})\in \B} \big[-k||\phi_{\omega}(s_t)-\phi_{\omega}(s_{t+1})||_2 
    - \log (
    1 + \sum_{s \in \B_{s \neq s_{t}}} f(s,s_{t+1}) )
     \big] 
     \label{eq:objectivelossb}
\end{align}

In the last line of \eqref{eq:objectivelossb}, we upper-bound $f_{\omega}(s_t,s_{t+1})$ with 1 since $e^{-v} < 1$ when $v$ is positive. The logarithmic function is monotonic, so the negative logarithm inverses the bound.

\section{Implementation details}\label{app:relabel}
In this section, we provide some details about \metname.

\paragraph{Number of negative samples.} In practice, to learn $\phi_{\omega}$ we do not consider the whole batch of states as negative samples. For each positive pair, we randomly sample only 10 states within the batch of states. This number has not been fine-tuned.

\paragraph{Relabeling strategies.}
We propose three relabeling strategies;
\begin{enumerate}

\item $\pi^{high}$ relabelling: we take samples from $\B^S$ and relabel the goal using clusters sampled with $\pi^{high}$ and randomly sampled states. This is interesting when the agent focuses on a task; this gives more importance to rewarding clusters and allows to forget uninteresting skills. We use it in \textit{Ant} and \textit{Half-Cheetah} environments.
\item Uniform relabelling: we take samples from $\B^S$ and relabel the goal using states sampled from from $\B^S$. When $\alpha_{skew} \approx 0$, this is equivalent to relabeling uniformly over the embedded state space. This is used for maze environments and \textit{Visual Door}.
\item Topological relabelling: we take samples from both $\B^S$ and $\B^G$ and relabel each goal with a state that belongs to a neighboring cluster. This is interesting when the topology is very large, making an uniform relabelling inefficient. This is applied in \textit{Visual Pusher} experiments, but we found it to work well in mazes and \textit{Visual Door} environments.
\end{enumerate}

\subsection{Comparison methods}

\paragraph{LESSON:} we used the code provided by the authors and reproduced some of their results in dense rewards settings. Since the environments are similar, we used the same hyper-parameter as in the original paper \cite{li2021learning}.

\paragraph{Skew-Fit:} since we use the same evaluation protocol, we directly used the results of the paper. In order to fairly compare DisTop to Skew-Fit, the state given to the DRL policy of DisTop is also the embedding of the true image. We do not do this in other environments. We also use the exact same convolutional neural network (CNN) architecture for weights $\omega$ as in the original paper of Skew-Fit. It results that our CNN is composed of three convolutional layers with kernel sizes: 5x5, 3x3, and 3x3; number of channels: 16, 32, and 64; strides: 3, 2 and 2. Finally, there is a last linear layer with neurons that corresponds to the topology dimensions $d$. This latent dimension is different from the ones of Skew-Fit, but this is not tuned and set to 10.

\paragraph{ELSIM:} we use the code provided by the authors. We noticed they used very small batch sizes and few updates, so we changed the hyper-parameters and get better results than in the paper on \textit{Half-Cheetah}. We set the batch size to $256$ and use neural networks with $2\times 256$ hidden layers. The weight decay of the discriminator is set to $1\cdot10^{-4}$ in the maze environment and $1\cdot10^{-3}$ in \textit{Ant} and \textit{Half-Cheetah}. In \textit{Ant} and \textit{Half-Cheetah}, the learning process was too slow since the agent sequentially runs up to 15 neural networks to compute the intrinsic reward; so we divided the number of updates by two. In our  results, it did not bring significant changes. 

\paragraph{SAC:} we made our own implementation of SAC. We made a hyper-parameter search on entropy scale, batch size and neural networks structure. Our results are consistent with the results from the original paper \cite{haarnoja2018soft}.

\section{Hyper-parameters}\label{app:hyperparam}

Table \ref{tab:hyperp} shows the hyper-parameters used in our main experiments. We emphasize that tasks are very heterogeneous and we did not try to homogenize hyper-parameters across environments.

\begin{table}[t]
    \centering
    \begin{tabular}{|c|c|c|}
        \hline
         Parameters & Values $RP/RD/MA/MC/SAM/SPM$ & Comments \\ \hline 
         DRL algorithm SAC & &  \\
         \hline
         Entropy scale & $0.1/0.1/0.2/0.2/0.1/0.2$ &  \\ 
         Hidden layers & $3/3/3/3/4/4$ &  \\ 
         Number of neurons & $512$ & Smaller may work\\ 
         Learning rate & RP: $3\cdot10^{-4}$ else $5\cdot10^{-4}$ & Works with both \\ 
         Batch size & RP: $256$ else $512$ &  Works with both  \\ 
         Smooth update & RP:$0.001$ else $0.005$ & Works with both. \\
         Discount factor $\gamma$ & $0.99/0.99/0.99/0.99/0.996/0.996$& Tuned for mazes \\
         \hline  
         Representation $\phi_{\omega}$ & & \\
         \hline
         Learns on $B^G$ & No/No/No/No/Yes/Yes & Works with both \\
         Learning rate & $1\cdot10^{-4}$, MA: $5\cdot10^{-4}$, MC: $1\cdot10^{-3}$ & Not tuned on MA, MC \\
         Number of neurons & 256 except robotic images & Not tuned \\
         Hidden layers & 2 except robotic images & Not tuned \\
         
         Distortion threshold $\delta$ & SPM: $0.01$ else $0.1$ & Tuned on SPM \\
         Distortion coefficient $k_c$ & 20 & See \appref{app:hyperparamstudy} \\
         Consistency coefficient $\beta$ & RD: $0.2$ else $2$ & Not tuned\\
         Smooth update $\alpha_{slow}$ & 0.001 & Not tuned \\
         Temperature $k$ & $1/1/3/3/3/3$ & See \appref{app:hyperparamstudy} \\
         Topology dimensions $d$ & $10/10/10/3/3/3$ & Not tuned \\ \hline
         
         OEGN and sampling& & \\ \hline
         Creation threshold $\delta_{new}$ & RP:$0.8$ else 0.6 & See \appref{app:hyperparamstudy} \\ 
         Success threshold $\delta_{success}$ & $\infty/\infty/0.2/0.2/\infty/\infty$ & \\
         Buffers size & $[8/15/5/5/15/15]\cdot10^{3}$ & \\
         Skew sampling $1+\alpha_{skew}$ & RD:$0.1$ else $0$ & See \appref{app:hyperparamstudy} \\ 
         updates per steps & $2/2/0.5/0.5/0.25/0.25$ &\\ \hline
         
         High-level policy $\pi^{high}$ & & \\ \hline
         Learning rate $\alpha_c$ & 0.05 & Tuned \\
         Neighbors learning rate & $0/0/0.2 \alpha_c /0.2 \alpha_c/0/0 $ & Not fine-tuned \\
         Skew selection $1+\alpha_{skew}'$ & $-1/-0,1/0/0/-1/-1/$ & See \appref{app:hyperparamstudy} \\
         Reward temperature $t^{ext}$ & $0/0/50/10/100/100$ & \\
         \hline
    \end{tabular}
    \caption{Hyper-parameters used in experiments.  RP, RD, MA, MC, SAM, SPM respectively stands for Robotic Visual Pusher, Robotic Visual Door, MuJoCo Ant, MuJoCo Half-Cheetah, Sparse Ant Maze, Sparse Point Maze.}
    \label{tab:hyperp}
\end{table}

\section{Environment details}\label{app:environment}

\subsection{Robotic environments}

Environments and protocols are as described in \cite{dblpPongDLNBL20}. For convenience, we sum up again some details here.
\paragraph{Visual Door:} a MuJoCo environment where a robotic arm must open a door placed on a table to a target angle. The state space is composed of 48x48 images and the action space is a move of the end effector (at the end of the arm) into (x,y,z) directions. Each direction ranges in the interval [-1,1]. The agent only resets during evaluation in a random state. During evaluation, goal-states are sampled from a set of images and given to the goal-conditioned policy. At the end of the 100-steps episode, we measure the distance between the final angle of the door and the angle of the door in the goal image.

\paragraph{Visual Pusher:} a MuJoCo environment where a robotic arm has to push a puck on a table. The state space is composed of 48x48 images and the action space is a move of the end effector (at the end of the arm) in (x,y) direction. Each direction ranges in the interval [-1,1]. The agent resets in a fixed state every 50 steps. During evaluation, goal-states are sampled randomly in the set of possible goals. At the end of the episode, we measure the distance between the final puck position and the puck position in the goal image.

\subsection{Maze environments}

These environments are described in \cite{nachum2018near} and we used the code modified by \cite{li2021learning}. For convenience, we provide again some details and explain our sparse version. The environment is composed of 8x8x8 fixed blocks that confine the agent in a U-shaped corridor displayed in \figureautorefname~\ref{fig:mazev1}. 

Similarly to \cite{li2021learning}, we zero-out the (x,y) coordinates and append a low-resolution top view of the maze to the proprioceptive state. This "image" is a 75-dimensional vector. In our sparse version, the agent gets 0 reward when its distance to the target position is below 1.5 and gets -1 reward otherwise. The fixed goal is set at the top-left part of the maze.

\paragraph{Sparse Point Maze:} the proprioceptive state is composed of 4 dimensions and its 2-dimensional action space ranges in the intervals [-1,1] for forward/backward movements and [-0.25,0.25] for rotation movements.

\paragraph{Sparse Ant Maze:} the proprioceptive state is composed of 27 dimensions and its 8-dimension action space ranges in the intervals [-16,16].

\section{Computational resources}\label{app:resources}

Each simulation runs on one GPU during 20 to 40 hours according to the environment. Here are the settings we used:
\begin{itemize}
    \item Nvidia Tesla K80, 4 CPU cores from of a Xeon E5-2640v3, 32G of RAM. 
    \item Nvidia Tesla V100, 4 CPU cores from a Xeon Silver 4114, 32G of RAM.
    \item Nvidia Tesla V100 SXM2, 4 CPU cores from a Intel Cascade Lake 6226 processors, 48G of RAM. (Least used).
\end{itemize}

\section{Example of skills}

\begin{figure}[t]
    \centering
    \includegraphics[width=0.7\linewidth]{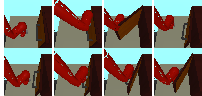}
    \caption{Examples of 8 skills learnt in \textit{Visual Door}.}
    \label{fig:doorexample}
\end{figure}

\begin{figure}[t]
    \centering
    \includegraphics[width=0.7\linewidth]{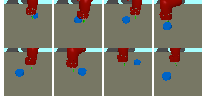}
    \caption{Examples of 8 skills learnt in \textit{Visual Pusher}.}
    \label{fig:pushexample}
\end{figure}

\begin{figure}[t]
    \centering
    \includegraphics[width=1\linewidth]{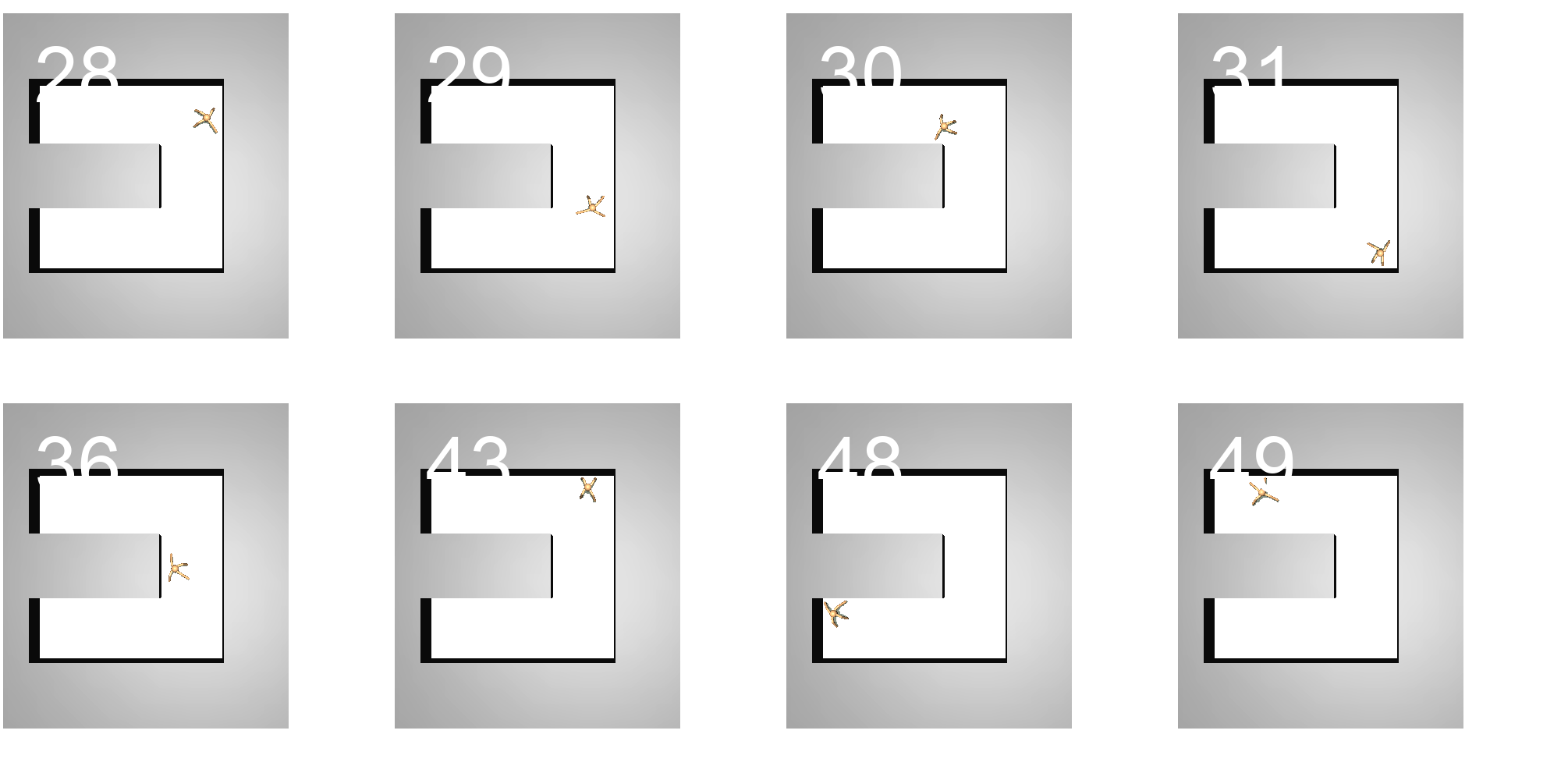}
    \caption{Examples of 8 skills learnt in \textit{Ant Maze}.}
    \label{fig:mazeexample}
\end{figure}

Figures \ref{fig:doorexample}, \ref{fig:pushexample}, \ref{fig:mazeexample} show examples skills learnt in respectively \textit{Visual Door}, \textit{Visual Pusher} and \textit{Ant Maze}. Additional videos of skills are available in supplementary materials. We also provide videos of the topology building process in maze environments. We only display it in maze environments since the 3D-topology is suitable.
\newpage 

\end{document}